\documentclass[letterpaper]{article} 
\usepackage{aaai24}  
\usepackage{times}  
\usepackage{helvet}  
\usepackage{courier}  
\usepackage[hyphens]{url}  
\usepackage{graphicx} 
\def\UrlFont{\rm}  
\usepackage{natbib}  
\usepackage{caption} 
\usepackage{algorithm}
\usepackage{algorithmic}

\usepackage{newfloat}
\usepackage{listings}
\DeclareCaptionStyle{ruled}{labelfont=normalfont,labelsep=colon,strut=off} 
\floatstyle{ruled}
\newfloat{listing}{tb}{lst}{}
\floatname{listing}{Listing}
\nocopyright 

\title{Closing the Gap in the Trade-off between Fair Representations and Accuracy }

\author {
    Biswajit Rout,
    Ananya B. Sai\thanks{presented at Deployable AI Workshop at AAAI-2024},
    Arun Rajkumar
}
\affiliations {
    
    Indian Institute of Technology Madras\\
    routb68@gmail.com, ananya@cse.iitm.ac.in, 
    arunr@cse.iitm.ac.in
}

\usepackage{bibentry}

\usepackage{amsmath}
\usepackage{subcaption}
\usepackage{tabularx}
\usepackage{booktabs}
\usepackage{devanagari}

\begin{document}

\maketitle

\begin{abstract}
The rapid developments of various machine learning models and their deployments in several applications has led to discussions around the importance of looking beyond the accuracies of these models. Fairness of such models is one such aspect that is deservedly gaining more attention. In this work, we analyse the natural language representations of documents and sentences (i.e., encodings) for any embedding-level bias that could potentially also affect the fairness of the downstream tasks that rely on them. 
We identify bias in these encodings either towards or against different sub-groups based on the difference in their reconstruction errors along various subsets of principal components. We explore and recommend ways to mitigate such bias in the encodings while also maintaining a decent accuracy in classification models that use them. 
\end{abstract}

\section{Introduction}
With the growing number of deployments of machine learning models in various fields and applications, it has become increasingly important to study the aspect of fairness of these models with respect to various features or sub-groups. For example, human face recognition applications need to work properly on all humans, irrespective of ethnicity, gender, age, and other features. Sentiment analysis models for social media should not favour or disfavor any demographic sub-groups or any topics in an unjustified manner. A credit scoring model, even when trained on historical biased data, should be carefully deployed so as to not perpetuate such biases, and base its decisions on the right factors. Similar requirements are applicable in various domains such as healthcare, justice system, education, and so on. 

In our work, we consider binary classification tasks, where one or more protected attributes are involved. That is, we can formalize our training data as consisting of $< X, P, Y >$, where $X$ is input data/ features, $Y$ is the binary label associated with $X$, that is to be predicted, and $P$ is a binary protected attribute which could be directly or indirectly present in $X$. When $P$ is not directly present, it is still possible for $P$ to potentially be inferred from the input. Whether directly present or inferred, the protected attribute(s) should ideally not influence the output $Y$. (There are also instances where knowledge of $P$ is useful / required for an ideal outcome \cite{10.1145/3287560.3287600_50_years_fairness, 10.5555/3295222.3295256_clustering_fairlets}. However, in our case, the discussions pertaining to tasks being agnostic to $P$ are more relevant.)

Depending on the task at hand, the irrelevance of $P$ on the outcome could be a strong preference (such as in hate-speech tagging) or an uncompromisable, strict requirement with serious implications (such as for determining credit card eligibility). In certain cases it could even be illegal to use $P$ within $X$ (such as for bail prediction).

This criteria of independence of $P$ has been well studied under the topic of fairness of machine learning models. However, in several cases it has been observed that the methods that improve fairness affect the performance of the models \cite{vu-etal-2020-multimodal,baldini-etal-2022-fairness_pretrained_models}. This can prove to be a hindrance to adopting such techniques of fairness in many applications. Our goal is to use fair representation or transformation of $X$ with respect to $P$ for downstream classification task such that the classifiers built on $X$ are less prone to bias towards specific subgroups or protected classes, while maintaining their overall accuracy.

We focus on classification problems in the Natural Language Processing (NLP) domain. The $X$ in our data are natural language sentences or documents. We consider 2 real world datasets: (i) Hindi Legal Document Corpus for predicting bail granting given the case files. The protected attribute in these cases is religion. (ii) English Twitter Corpus for hate speech tagging of tweets, in which the protected attributes are ethnicity and gender. We encode the Hindi documents using FastText embeddings \cite{kumar-etal-2020-passage_hindi_fasttext} and the English sentences using GloVe embeddings. We compute the sentence / document-level encoding of these sentences using 2 different approaches: vector average \cite{landauer1997solution_vector_averaging1} and vector extrema \cite{vectorextrema}. We feed these encodings to SVM to perform the binary classification task. 

To measure the fairness of these encodings, we analyse them
for any differences in their preciseness in representing each of the sub-groups of protected attributes. Specifically we examine the differences in the reconstruction errors of various groups along the initial principle components. 
We show that both the strategies of vector average and vector extrema show some bias towards particular sub-groups in different scenarios / datasets. 
We also notice a trade-off with the fairer approach usually being less accurate than the one containing the bias. 
To bring out the best of both accuracy and fairness, we propose using a convex combination of both the approaches. 
We provide recommendations for choosing an optimal combination based on the available leeway to compromise on accuracy traded-off with the strictness of the requirement for representation-level fairness in the classification tasks.
\section{Related Work}
\label{sec:related_work}

The studies on fairness of machine learning models are of various forms (eg., algorithmic fairness, demographic fairness, etc) and at various stages of model-building (such as data-level, representation-level, outcome-level, etc). The fairness studies can also be categorised based on the end-goal or the downstream task they address. We position our work as a combination of representation-level fairness analysed in the context of binary classification tasks. 


In the following two paragraphs of related work, we first review fairness studies on the classification tasks. (Note the distinction between classification tasks that have a label different from the protected attribute(s) and the classification tasks that are set up to predict the protected attribute. Both of these task types are relevant to the topic of fairness in different ways. We however discuss works of the former type here due to its relevance with the setting of our current work.) We then review the works that focus on fairness at the intermediate data representation-level.
To the best of our knowledge, the particular combination of investigating the principle components of natural language data representations for a classification task that needs to be agnostic to a set of protected attributes has not been studied so far.

\citet{prost-etal-2019-debiasing_bios} have a multi-task classification task to predict the occupation based on biographies of a person, with gender as the protected attribute. They propose to use `strongly debiased embeddings' where all the embeddings are projected into a subspace orthogonal to the gender subspace. They measure fairness at the outcome level using `Equality of Opportunity' \cite{hardt2016equality_opportunity}, where the true positive rate should be independent of the protected attribute value.
\citet{huang-2022-easy_multilingual_classification} presents a domain adaptation approach to improve fairness of classifiers in multilingual settings without compromising too much on performance. They do so by augmenting the training data to include domain features (in this case, the protected attributes) and use multiple domain-dependent feature extractors and one domain-independent one. At the time of testing, only the domain-independent features are used. They measure performance using F1-macro score and area under ROC curve, and measure fairness by using the `equality differences' \cite{10.1145/3278721.3278729_measuring_bias_text_classification} of false positive / negative rates. Various post-processing methods to improve fairness using pretrained language models were explored by \citet{baldini-etal-2022-fairness_pretrained_models} in the toxic text classification task with religion, race and gender as the protected attributes.

One of the earliest methods for understanding the representation-level bias in NLP in the realm of embeddings was the `Word-Embedding Association Test' (WEAT) \cite{doi:10.1126/science.aal4230_weat}. This measures the association of regular words with the words corresponding to the protected attributes based on the cosine similarities of their GloVe embeddings. The authors show that the embeddings reflect both harmless and harmful biases held in society. \citet{may-etal-2019-measuring_seat} extend this work to analyse sentence-level encodings by proposing `Sentence Encoder Association Test' (SEAT). In our work, we instead develop a method to analyse encodings based on Principle Component Analysis (PCA), which is commonly used to reduce the dimensionality of word embeddings. It addresses the question of how recoverable are the encodings of sentences / documents pertaining to different values of protected attributes.
The closest work to ours is by \citet{10.5555/3327546.3327755_price_of_fair_pca}, which is outside the NLP domain. They introduce the notion of Fair PCA in the context of images and structured / tabular data. They propose an algorithm to perform a more fair dimensionality reduction, given the representations of the data (which are considered fixed). In our work, we look at the problems in NLP and propose ways to find more fair representations of the data instead (i.e., the PCA / the dimensionality reduction method is fixed), while also maintaining a desired accuracy.


\section{Experimental Setup}

\subsection{Datasets}
For our experiments, we use datasets that have a protected attribute along with a classification task that should ideally be independent of that attribute. We consider 2 such datasets as described below:
\\
(i) \textbf{HLDC (Hindi Legal Document Corpus)} for Bail Prediction \cite{kapoor-etal-2022-hldc} - This dataset contains case documents in Hindi language from district courts in the Indian state of Uttar Pradesh between the period May 01, 2019 to May 01, 2021. It contains documents of the case facts and arguments, judge summary, and the case results. 
From manual inspection of the metadata of the case files the authors have categorized the cases into 300 unique types of cases out of which bail applications have the highest percentage (31.71\%) of presence. For our task we use only the bail application cases with religion as the protected attribute (also, we consider only Hindu and Muslim religions in this study). We consider two segments of case files viz., the facts \& arguments and the case results . We do not consider the judge summaries in the files since they may contain information regarding the bail outcomes.
\\
(ii) \textbf{MTC (Multilingual Twitter Corpus)} for Hate Speech Recognition \cite{huang-etal-2020-multilingual_mtc_hate_recog} - We consider the English subset of this dataset which contains anonymized Twitter posts that have been labelled as hate-speech or not. While the dataset is anonymized, certain features / attributes of the users are made available, including gender, race, and age of the user. The values of each attribute are converted to a binary format by the authors as follows: male / female for gender,  white / non-white for race. For age, the median age is computed and categorised as above median or below / equal to median. 
We consider the samples containing non-null values for ethnicity and gender.

\subsection{Data Preprocessing} 
The HLDC dataset authors release the data after Optical Character Recognition (OCR) and Named-Entity removal (NER). 
That is, after extracting the case files from ecourt-website in raw pdf format with Devanagari script, they perform OCR via Tesseract tool which reportedly works robustly here since most of the files are well typed documents. They have anonymized the location names, first names, middle names, and last names by replacing them with {\dn nAm} (`Naam' or name). They used a gazetteer along with regex based rules for NER to anonymize the data. They also run a RNN-based Hindi NER model to find additional named entities and subsequently anonymized them. Judge names are also anonymized since they can be correlated with outcomes of the case. 
Following \citet{girhepuje2023models}, we categorize the cases into Hindu or Muslim by using a subset of names commonly found in these communities that are still present in the case files even after the above preprocessing steps by the HLDC authors. 
We perform further preprocessing where we removed special characters (such as copyright symbols) by checking if the characters lie outside a desired range of ASCII values. 
We found some cases had length less than 30 words after preprocessing, which is much lesser than the average document length of 187.2 words. We drop out such cases.

For the MTC dataset, the authors have released anonymized tweets by replacing all usernames by `USER'. Additionally all hyperlinks and hashtags were replaced by `URL' and `HASHTAG' respectively. For tokenization of tweets, we use the NLTK tokenizer. 

Since these datasets have different number of samples with each attribute value and label value, we decide to synthetically balance the dataset by sampling equal number of data points from each attribute value of interest.
The statistics of these datasets are summarized in Tables~\ref{tab:hdlc_data_stats} and \ref{tab:mtc_data_stats}.
Here onwards, we refer to the gender-balanced sampled subset of the MTC dataset as MTC-gen and the ethnicity-based sampled subset as MTC-eth. HDLC dataset from here on refers to our religion-balanced sampled subset. Note that we refer to each value of a protected attribute as a `group'.
\begin{table}[!ht]
    \centering
    \begin{tabular}{c |c|c| c| c | c }
        Group & num & avg & min & max & \% bail  \\
        \hline
        Total & 1192 & 187.26 & 30 & 537 & 41.86 \\
        Hindu & 596 & 189.58 & 37 & 537 & 45.13\\
        Muslim & 596 & 184.62 & 30 & 491 & 38.59  
    \end{tabular}
    \caption{HLDC Dataset statistics showing number of samples (num), average(avg), minimum and maximum number of words per sample, and the percentage (\% bail) of samples with bail granted label.}
    \label{tab:hdlc_data_stats}
\end{table}
\begin{table}[!ht]
    \centering
    \begin{tabular}{c |c|c| c| c | c}
        Group & num & avg & min & max & \% hate  \\
        \hline
        MTC-gen & 4000 & 17.28 & 10 & 40 & 36.08 \\ 
        MTC-eth & 4000 & 17.28 & 10 & 40 & 38.65
    \end{tabular}
    \caption{Dataset statistics for MTC-gen and MTC-eth showing number of samples, average, minimum and maximum number of words per sample, and the percentage of samples that are labelled hate-speech}
    \label{tab:mtc_data_stats}
\end{table}

\begin{figure*}[!ht]
\begin{subfigure}{.49\textwidth}
    \centering
    \includegraphics[width=1\columnwidth]{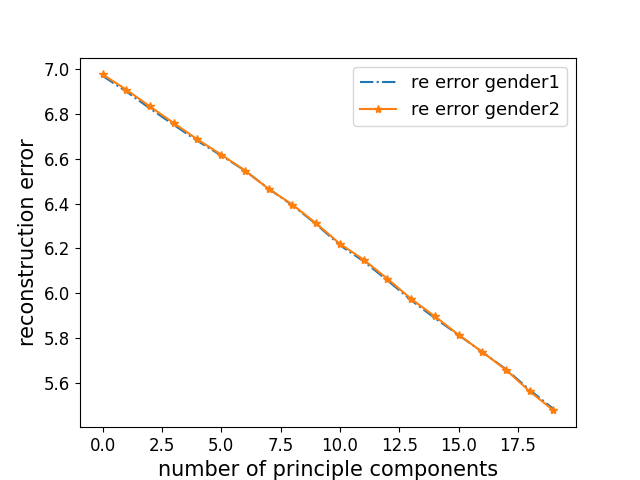}
    \caption{MTC-gen with Vector Extrema}
    \label{fig:mtcg_ve}
\end{subfigure}
\begin{subfigure}{.49\textwidth}
    \centering
    \includegraphics[width=1\columnwidth]{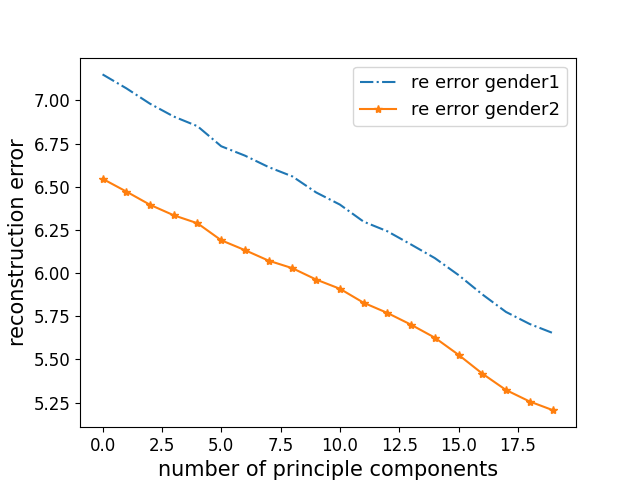}
    \caption{MTC-gen with Vector Averaging}
    \label{fig:mtcg_va}
\end{subfigure}
\begin{subfigure}{.49\textwidth}
    \centering
    \includegraphics[width=1\columnwidth]{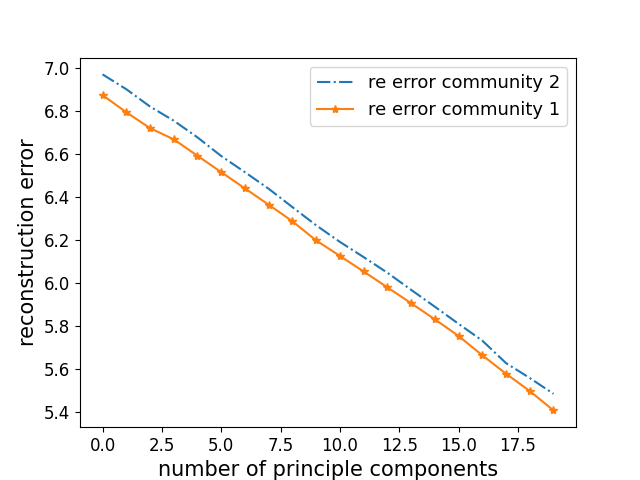}
    \caption{HLDC with Vector Extrema}
    \label{fig:hldc_ve}
\end{subfigure}
\begin{subfigure}{.49\textwidth}
    \centering
    \includegraphics[width=1\columnwidth]{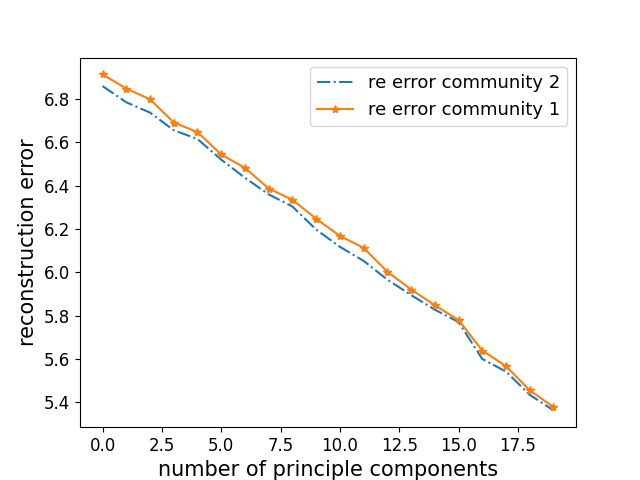}
    \caption{HLDC with Vector Averaging}
    \label{fig:hldc_va}
\end{subfigure}
\caption{PCA reconstruction error for MTC-gen and HLDC datasets at each number of dimensions used for PCA reconstruction with each of vector extrema and vector averaging techniques}
\label{fig:pca_re}
\end{figure*}
\begin{figure*}[!ht]
\begin{subfigure}{.49\textwidth}
    \centering
    \includegraphics[width=1\columnwidth]{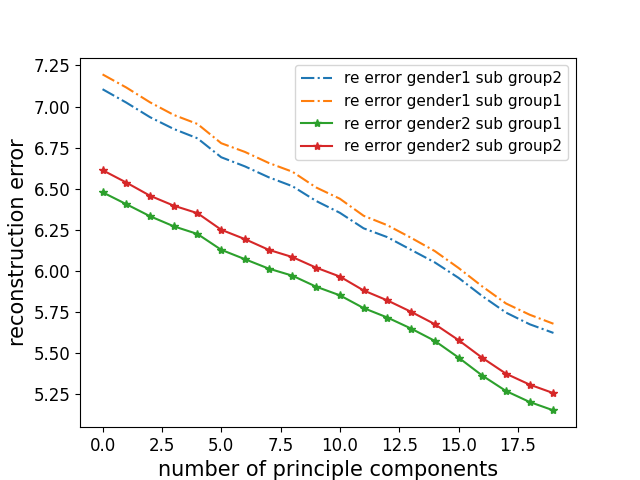}
    \caption{MTC-gen with vector average}
    \label{fig:mtcg_split_va}
\end{subfigure}
\begin{subfigure}{.49\textwidth}
    \centering
    \includegraphics[width=1\columnwidth]{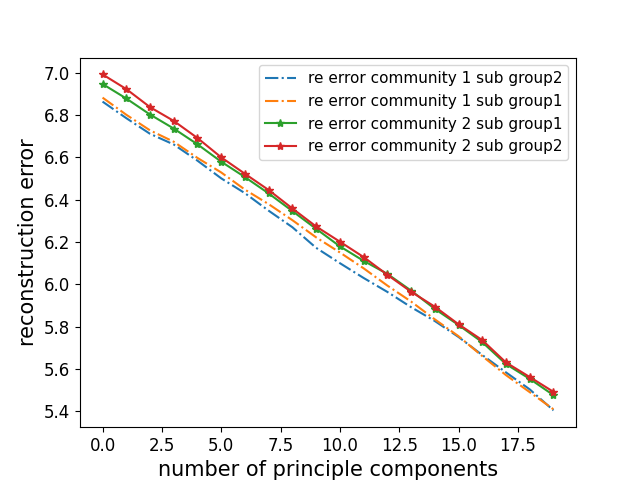}
    \caption{HLDC with vector extrema}
    \label{fig:hldc_split_ve}
\end{subfigure}
\caption{PCA Reconstruction Error within groups (i.e., random sub-groups of same group) and across groups}
\label{fig:pca_rand_re}
\end{figure*}

\subsection{Word embeddings}
For Hindi word embeddings, we use the FastText embeddings released by \citet{kumar-etal-2020-passage_hindi_fasttext} of 50 dimensions. These allow for unseen words to also be assigned embeddings based on sub-words / characters. 
For English word embeddings, we use the 50 dimensionsal GloVe embeddings which are pretrained on Twitter data.
The following sub-sections discuss some of the simple strategies used by the NLP community for combining the word embeddings into a representation for the sentence or document.

\subsection{Vector Averaging}
One of the simplest ways to aggregate word embeddings to a sentence-level or document-level representation is to average the embeddings of all the words in the sentence / document. The Vector Averaging technique \cite{landauer1997solution_vector_averaging1} does exactly this by computing a sentence-level encoding, $\overrightarrow{s}$, by averaging the word embeddings $\overrightarrow{w}$ of all the tokens $w$ in the sentence, $s$.
$$  \overrightarrow{s} = \frac{\sum_{w\in s}\overrightarrow{w}}{|s|} $$


\subsection{Vector Extrema}
Another approach to compute sentence-level encodings is to use Vector Extrema \cite{vectorextrema}. In this case, a $k$-dimensional sentence representation is constructed by considering the dimension-wise extreme-value among the $k$-dimensional word embeddings of all the words in the sentence. The most extreme value along each dimension is the value farthest from $0$ and can be defined as follows:
\begin{align*}
    \overrightarrow{s_d} = 
    \begin{cases}
     \max_{w \in s}\overrightarrow{w}_d, & \text{if } \overrightarrow{w}_d>|\min_{w'\in s}\overrightarrow{w'}_d| \\
     \min_{w\in s}\overrightarrow{w}_d & \text{otherwise}
    \end{cases}
\end{align*}

where $\overrightarrow{s_d}$ is the $d^{\text{th}}$ dimension of the sentence encoding $\overrightarrow{s}$. Similarly, $\overrightarrow{w_d}$ is the $d^{\text{th}}$ dimension of the word embeddings.


\subsection{Models and Performance Measures}
\label{sec:models_perf_measures}
We train Support Vector Machine (SVM) models using each of the sentence-encoding strategies to perform classification on the 2 datasets considered. 
We split the dataset into $\sim80\%$ training data and $\sim 20\%$ test data. 
We use SVM models 
with the RBF (Radial Basis Function) kernel. To fit the parameters `C' and `$\gamma$', we use 2 approaches: For HLDC, which has fewer samples of data (i.e., 1192 data points), we perform 5-fold cross validation with grid search on 1000 data points to find the best performing `C' and `$\gamma$'. The test accuracies reported in our paper are on the 192 test data points.
For MTC-gen and MTC-eth, we divide the 80\% of the initial 80-20 train-test split further into a 80-20 split to use as training data and validation data.
We perform a coarse grid search for hyper-parameters `C' and `$\gamma$' by varying the values exponentially in the RBF-kernel based SVM. We used LIBSVM for training \cite{10.1145/1961189.1961199_libsvm}. We repeat the same process for each sentence encoding strategy adopted or proposed in this paper. We measure the performance of these models based on their accuracy on the classification task. 
We discuss our approaches for measuring / understanding the fairness of these models in section~\ref{sec:fairness}.



\section{Performance on Classification Task}
We first measure the performance on classification task when using the simple strategies of aggregating word embeddings with vector averaging and vector extrema. We use each of these approaches to encode the documents (in HLDC) and sentences (in MTC-gen and MTC-eth) and use that as input to SVM classifiers. We train on ~80\% of the data (creating any validation data out of this as required) and test on the remaining. Table~\ref{tab:svm_accuracies} shows the test accuracies obtained by fitting SVM models as per the methods outlined in section~\ref{sec:models_perf_measures}.
We find a noticeable difference in performance on the datasets when each of vector extrema and vector average are used. Vector-average consistently performs better than vector extrema on all the datasets. 
\begin{table}[!ht]
    \centering
    \begin{tabular}{c|c|c}
        Dataset & Encoding & Accuracy \\
        \hline
        MTC-gen & extrema & 63.00\%  \\
        MTC-gen & average & 79.63\% \\
        MTC-eth & extrema & 62.00\% \\
        MTC-eth & average & 84.38\% \\
        HLDC & extrema & 65.97\% \\
        HLDC & average & 74.35\% 
    \end{tabular}
    \caption{Classification Accuracies of SVM on HLDC and MTC datasets using vector-extrema and vector-average techniques }
    \label{tab:svm_accuracies}
\end{table}


Apart from the performance, we also want to understand how each of these approaches differ in the fairness aspect. We discuss this further in the next section. 

\section{Fairness of Representations}
\label{sec:fairness}
Understanding fairness at the representation-level is important and can have cascading effects on various downstream tasks and applications \cite{10.5555/3327546.3327755_price_of_fair_pca,may-etal-2019-measuring_seat}.  As noted in section~\ref{sec:related_work}, the most common approaches to study representation-level fairness in NLP are based on the distance / similarity of encodings with the encodings of the protected features.
Here, we propose an alternate approach that focuses on whether various encoding strategies represent data of different groups to a similar degree. 
Specifically, we study whether
the major principle components of the data capture unbalanced amounts of information about each group or differentiate data representations based on the protected attribute. To measure this, we perform Principle Component Analysis (PCA) on the whole balanced datasets and compute the reconstruction error of each group separately, for various number of reduced dimensions. We then evaluate whether or not the reconstruction error is similar for all groups. If it varies, we also want to determine by how much and put that quantity into perspective for a better understanding and drawing reasonable conclusions.

\subsection{PCA Reconstruction Errors per Group}
Figure~\ref{fig:pca_re} shows the reconstruction error on documents / sentences pertaining to different groups. (Due to limited space, we show the reconstruction error plots for MTC-ethnicity in appendix since the results are very similar to the MTC-gender plots). We notice for both the MTC-gen and MTC-eth datasets, vector extrema has the same reconstruction error for all the gender and ethnicity groups. As seen in Figure~\ref{fig:mtcg_ve}, this holds true for reconstructions using different number of dimensions / principle components. However, vector average shows a clear difference in the reconstruction errors based on ethnicity and gender. As seen in Figure~\ref{fig:mtcg_va}, gender 2 consistently has lesser reconstruction error than gender 1 across experiments with different numbers of principle components.
In case of HLDC, we find that both approaches have non-negligible differences in reconstruction errors of different communities. However, the difference in vector extrema is wider than vector average and more consistently present across various dimension numbers used in PCA. This can be observed in Figures~\ref{fig:hldc_ve} and \ref{fig:hldc_va}. We also find that community 1 has lesser reconstruction error when vector extrema is used, whereas community 2 has lesser reconstruction error on average when vector average is used.  (In Figures~\ref{fig:hldc_ve} and \ref{fig:hldc_va}, community 1 refers to Muslims and community 2 is Hindus)

\begin{figure*}[!ht]
\begin{subfigure}{.49\textwidth}
    \centering
    \includegraphics[width=1\columnwidth]{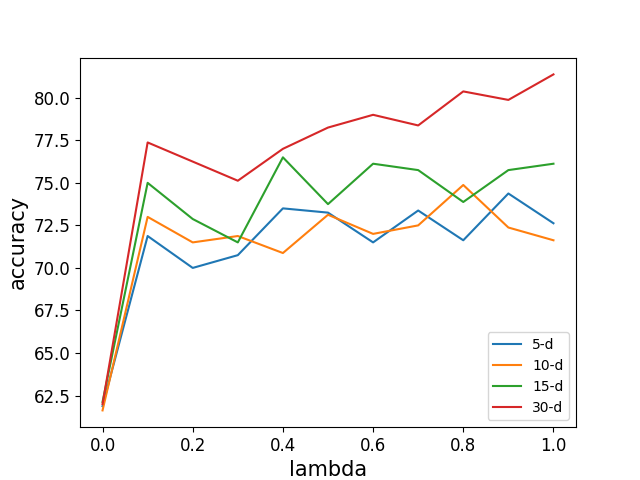}
    \caption{SVM accuracies on MTC-eth dataset}
    \label{fig:gen_svm1}
\end{subfigure}
\begin{subfigure}{.49\textwidth}
    \centering
    \includegraphics[width=1\columnwidth]{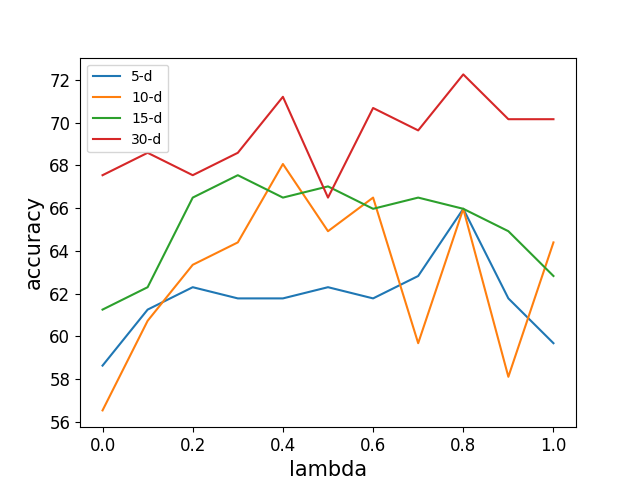}
    \caption{SVM accuracies on HLDC dataset}
    \label{fig:gen_svm3}
\end{subfigure}
\caption{Accuracy of SVMs while using different number of dimensions (5d, 10d, 15d, 30d) at various values of $\lambda $ to form a convex combination of vector average and vector extrema encodings}
\label{fig:svm_accuracies_lambda}
\end{figure*}

\begin{figure*}[!ht]
\begin{subfigure}{.33\textwidth}
    \centering
    \includegraphics[width=1\columnwidth]{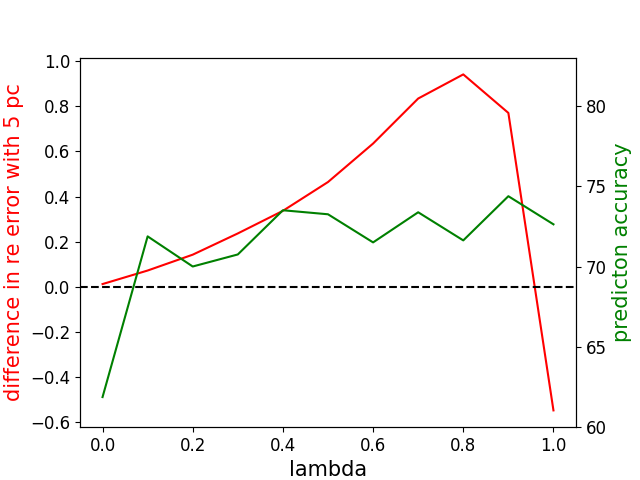}
    \caption{MTC-eth with 5 dim PCA}
    \label{fig:eth_pd1}
\end{subfigure}
\begin{subfigure}{.33\textwidth}
    \centering
    \includegraphics[width=1\columnwidth]{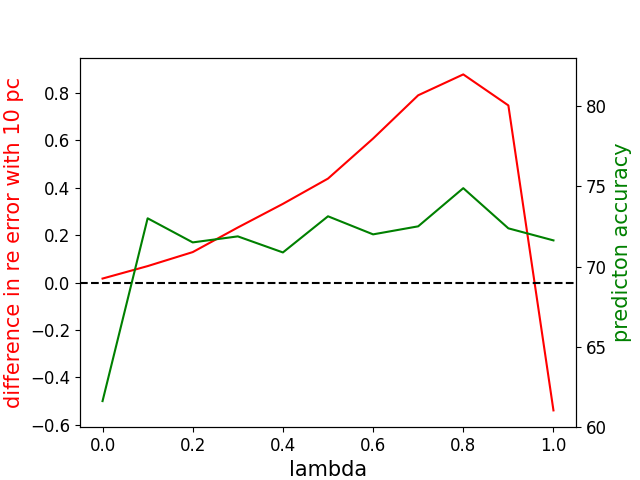}
    \caption{MTC-eth with 10 dim PCA}
    \label{fig:eth_pd2}
\end{subfigure}
\begin{subfigure}{.33\textwidth}
    \centering
    \includegraphics[width=1\columnwidth]{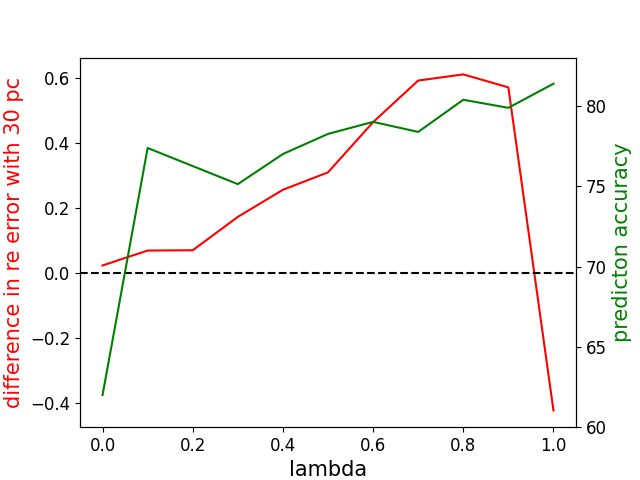}
    \caption{MTC-eth with 30 dim PCA}
    \label{fig:eth_pd3}
\end{subfigure}
\begin{subfigure}{.33\textwidth}
    \centering
    \includegraphics[width=1\columnwidth]{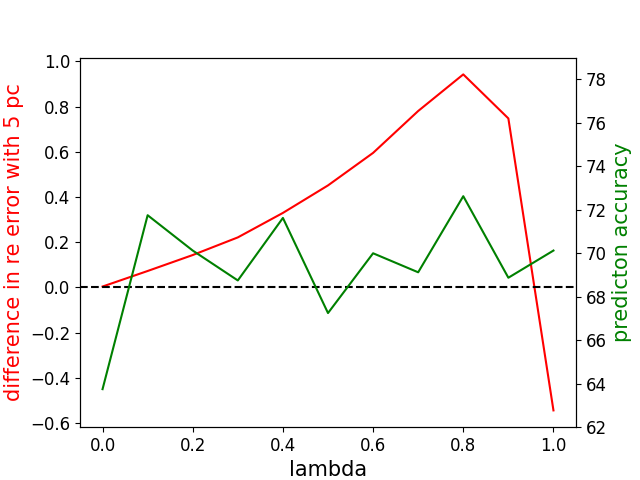}
    \caption{MTC-gen with 5 dim PCA}
    \label{fig:gen_pd1}
\end{subfigure}
\begin{subfigure}{.33\textwidth}
    \centering
    \includegraphics[width=1\columnwidth]{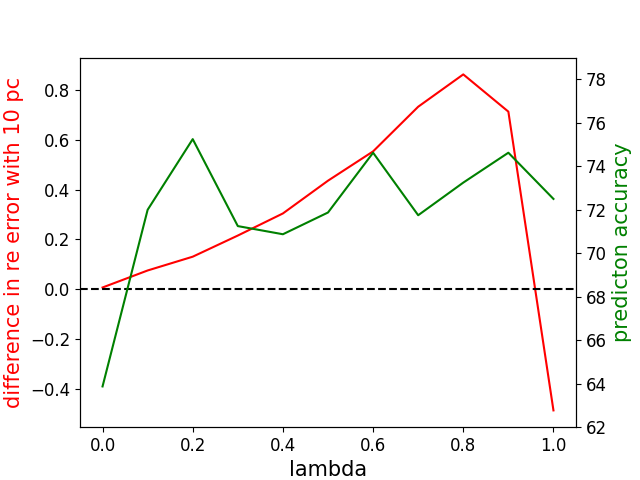}
    \caption{MTC-gen with 10 dim PCA}
    \label{fig:gen_pd2}
\end{subfigure}
\begin{subfigure}{.33\textwidth}
    \centering
    \includegraphics[width=1\columnwidth]{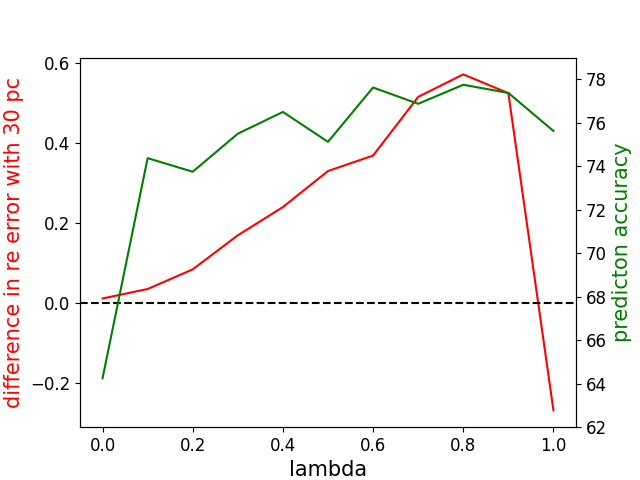}
    \caption{MTC-gen with 30 dim PCA}
    \label{fig:gen_pd3}
\end{subfigure}
\begin{subfigure}{.33\textwidth}
    \centering
    \includegraphics[width=1\columnwidth]{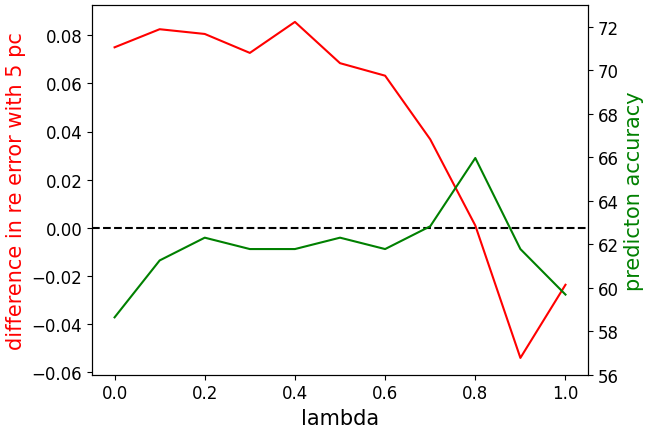}
    \caption{HLDC with 5 dim PCA}
    \label{fig:hdlc_pd1}
\end{subfigure}
\begin{subfigure}{.33\textwidth}
    \centering
    \includegraphics[width=1\columnwidth]{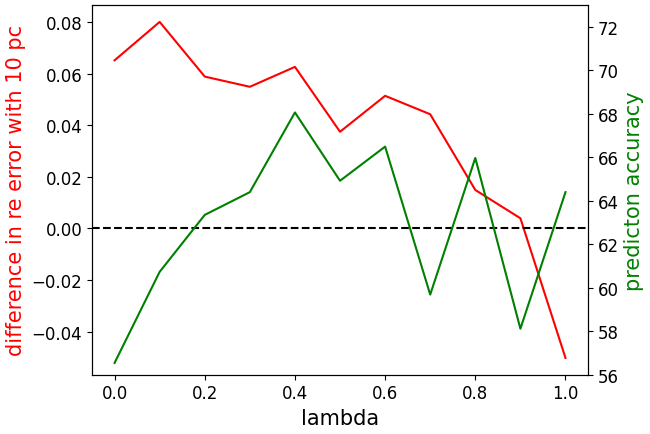}
    \caption{HLDC with 10 dim PCA}
    \label{fig:hldc_pd2}
\end{subfigure}
\begin{subfigure}{.33\textwidth}
    \centering
    \includegraphics[width=1\columnwidth]{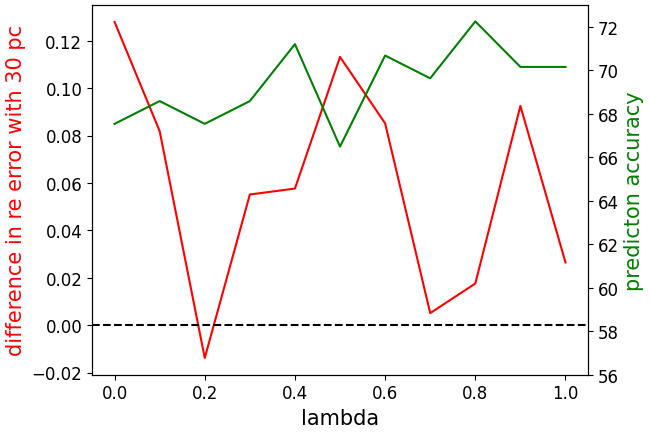}
    \caption{HLDC with 30 dim PCA}
    \label{fig:hldc_pd3}
\end{subfigure}
\caption{PCA reconstruction difference versus classification accuracy for various  $\lambda $ convex combination of vector average and vector extrema, at various dimensions of PCA reconstruction}
\label{fig:tradeoff}
\end{figure*}
\subsection{Comparison with random splits}
We want to investigate whether the gap in the reconstruction errors can occur by chance and benchmark the gap. Towards this end, we analyse the PCA reconstruction errors on splitting data from the same group further into randomly assigned sub-groups. This would contrast the inter-group error differences with the intra-group error differences. This is shown in the plots of Figure~\ref{fig:pca_rand_re} for vector-extrema for HLDC and vector-average for MTC-gen.
We find around 10x difference between the gaps in the random splits within group and the gaps in splits across groups of gender in the MTC-gen dataset when using vector average. Refer Figure~\ref{fig:mtcg_split_va}. The difference is not as pronounced, but still discernible for the HLDC dataset with vector extrema. When vector extrema is used for MTC-gen, the inter-group and intra-group error differences are comparable. This is presented in the appendix for contrasting with the other plots. (We find the MTC-eth plots to be very similar to reconstruction error plots for MTC-gen, so the same observations hold for MTC-eth). 

In order to further understand how much of the difference in reconstruction errors comes from the magnitude of vectors, we compute the length of the vectors, i.e., the Frobenius norm of the encodings for each encoding strategy. We report the results in Table~\ref{tab:vector_lengths}. Since the differences in the vector lengths across groups are well within the standard deviations of lengths of vectors within the same group, we conclude that these are not significant differences. Further exploration and analysis of the vector directions could potentially be beneficial than magnitudes of vectors.

\begin{table}[!ht]
    \centering
    \begin{tabular}{c|c|c|c|c}
        Data & len(VE)  & sd(VE) & len(VA) & sd(VA) \\
        \hline
        MTC-gen & 10.25 & 0.54 & 4.78 & 0.45 \\
        Gen 1 & 10.28 & 0.53 & 4.86 & 0.42 \\
        Gen 2 & 10.22 & 0.55 & 4.70 & 0.47 \\
    \end{tabular}
    \caption{Vector lengths and standard deviations}
    \label{tab:vector_lengths}
\end{table}

\section{Improving Fairness and Maintaining Performance}
From the previous sections, we observe that vector-average has better accuracies on MTC datasets, but is more unfair with respect to the PCA reconstruction errors. Vector-extrema shows a complementary behaviour to this on both the classification performance and fairness. On the other hand, in case of HLDC dataset, we do not find any such explicit trend, but observe the scope for more balance. 
Our aim is to improve fairness without sacrificing the accuracy / performance by much. 
Towards this end, we first explore if there exists a convex combination of the two approaches that can give us the desired result. 
\subsection{Convex Combination of Embeddings}
We experiment with the following formulation for combining the vector average encoding, $\overrightarrow{avg}$ and vector extrema encoding $\overrightarrow{ext}$:
\begin{align}
    \overrightarrow{s} = \lambda \ \overrightarrow{avg} + (1-\lambda) \  \overrightarrow{ext}
\end{align}
where $\lambda $ is varied from $0.0$ to $1.0$ with a step-size of $0.1$.

At each of these $\lambda$ values, we use the resulting combined sentence encodings to train / fit SVM classifiers for the corresponding binary classification task. 
At each value of $\lambda$, we find the best performing C and $\gamma$ hyper-parameters using the methods described in section \ref{sec:models_perf_measures}. Using these, we compute the accuracy on the test data. We repeat this for various number of dimensions of PCA reconstruction of the MTC-eth, MTC-gen, and HLDC dataset. The accuracies for 5, 10, 15, and 30 dimensional reconstructions of the MTC-eth and HLDC datasets are shown in Figure~\ref{fig:svm_accuracies_lambda} while using various values of $\lambda$ to combine vector-average and extrema encodings.
(A similar plot for MTC-gen is presented in appendix.) As one might expect, the accuracies tend to improve as the number of dimensions used increase. For MTC-eth, the accuracies are higher on the vector-average end of the graph (i.e., $\lambda \to 1 $). For HLDC, where vector-average and extrema have relatively similar performance, the accuracies are higher around some of the intermediate values of $\lambda$. 


For each $\lambda$ value, we also compute the difference in the PCA reconstruction error for each group. At an optimal $\lambda$, the difference in the errors should be $0$ for all subsets of dimensions or principle components considered. On the other hand, we also want the optimal $\lambda$ value to yield a combination of the encodings that is within an acceptable $\epsilon$ difference in accuracy.
We plot the PCA reconstruction error as well as the classification accuracy of the SVM model trained on the $\lambda$-combined encodings for each $\lambda$ . 
This is presented in Figure~\ref{fig:tradeoff} for various dimensions of PCA reconstructions of the HLDC data, MTC-gen and MTC-eth data. The red curves and the left side y-axis of the plot correspond to the error difference. The dashed line represents $0$ difference in reconstruction error. The green curve and the right side labels of y-axis correspond to the classification accuracies.
We identify the ideal values of $\lambda = $ 0.97 for the both the variants of MTC dataset, which reduces the difference in reconstruction error, while also consistently maintaining the accuracy. 
The recommendations for HLDC are dimension-dependent and can be derived by fixing an acceptable range of error-difference and maximising the accuracy in that range. 

\section{Conclusion and Future Directions}
We present an approach to examine the representation-level bias by means of analysing differences in PCA reconstruction errors of various groups. We show that a simple convex combination of vector extrema and vector average can mitigate this form of representational bias while maintaining the accuracy within a reasonable range.

As part of future work, we plan to find methods beyond grid search to find the optimal $\lambda$. We also want to further understand if the encodings with the desired properties can be obtained through direct training instead of the heuristic combination approach explored here. We aim to better understand the influences of this form of representation-fairness on subsequent steps, such as the classifier. We believe representation-level fairness would be relevant to study in tasks other than classification and hope future works explore the applicability of our approach in other applications.





\bibliography{aaai24}

\begin{thebibliography}{18}
\providecommand{\natexlab}[1]{#1}

\bibitem[{Baldini et~al.(2022)Baldini, Wei, Natesan~Ramamurthy, Singh, and Yurochkin}]{baldini-etal-2022-fairness_pretrained_models}
Baldini, I.; Wei, D.; Natesan~Ramamurthy, K.; Singh, M.; and Yurochkin, M. 2022.
\newblock Your fairness may vary: Pretrained language model fairness in toxic text classification.
\newblock In Muresan, S.; Nakov, P.; and Villavicencio, A., eds., \emph{Findings of the Association for Computational Linguistics: ACL 2022}, 2245--2262. Dublin, Ireland: Association for Computational Linguistics.

\bibitem[{Caliskan, Bryson, and Narayanan(2017)}]{doi:10.1126/science.aal4230_weat}
Caliskan, A.; Bryson, J.~J.; and Narayanan, A. 2017.
\newblock Semantics derived automatically from language corpora contain human-like biases.
\newblock \emph{Science}, 356(6334): 183--186.

\bibitem[{Chang and Lin(2011)}]{10.1145/1961189.1961199_libsvm}
Chang, C.-C.; and Lin, C.-J. 2011.
\newblock LIBSVM: A Library for Support Vector Machines.
\newblock \emph{ACM Trans. Intell. Syst. Technol.}, 2(3).

\bibitem[{Chierichetti et~al.(2017)Chierichetti, Kumar, Lattanzi, and Vassilvitskii}]{10.5555/3295222.3295256_clustering_fairlets}
Chierichetti, F.; Kumar, R.; Lattanzi, S.; and Vassilvitskii, S. 2017.
\newblock Fair Clustering through Fairlets.
\newblock In \emph{Proceedings of the 31st International Conference on Neural Information Processing Systems}, NIPS'17, 5036–5044. Red Hook, NY, USA: Curran Associates Inc.
\newblock ISBN 9781510860964.

\bibitem[{Dixon et~al.(2018)Dixon, Li, Sorensen, Thain, and Vasserman}]{10.1145/3278721.3278729_measuring_bias_text_classification}
Dixon, L.; Li, J.; Sorensen, J.; Thain, N.; and Vasserman, L. 2018.
\newblock Measuring and Mitigating Unintended Bias in Text Classification.
\newblock In \emph{Proceedings of the 2018 AAAI/ACM Conference on AI, Ethics, and Society}, AIES '18, 67–73. New York, NY, USA: Association for Computing Machinery.
\newblock ISBN 9781450360128.

\bibitem[{Forgues et~al.(2014)Forgues, Pineau, Larchev{\^e}que, and Tremblay}]{vectorextrema}
Forgues, G.; Pineau, J.; Larchev{\^e}que, J.-M.; and Tremblay, R. 2014.
\newblock Bootstrapping Dialog Systems with Word Embeddings.
\newblock In \emph{NeurIPS, modern machine learning and natural language processing workshop}, volume~2.

\bibitem[{Girhepuje et~al.(2023)Girhepuje, Goel, Krishnan, Goyal, Pandey, Kumaraguru, and Ravindran}]{girhepuje2023models}
Girhepuje, S.; Goel, A.; Krishnan, G.~S.; Goyal, S.; Pandey, S.; Kumaraguru, P.; and Ravindran, B. 2023.
\newblock Are Models Trained on Indian Legal Data Fair?
\newblock arXiv:2303.07247.

\bibitem[{Hardt, Price, and Srebro(2016)}]{hardt2016equality_opportunity}
Hardt, M.; Price, E.; and Srebro, N. 2016.
\newblock Equality of opportunity in supervised learning.
\newblock \emph{Advances in neural information processing systems}, 29.

\bibitem[{Huang(2022)}]{huang-2022-easy_multilingual_classification}
Huang, X. 2022.
\newblock Easy Adaptation to Mitigate Gender Bias in Multilingual Text Classification.
\newblock In Carpuat, M.; de~Marneffe, M.-C.; and Meza~Ruiz, I.~V., eds., \emph{Proceedings of the 2022 Conference of the North American Chapter of the Association for Computational Linguistics: Human Language Technologies}, 717--723. Seattle, United States: Association for Computational Linguistics.

\bibitem[{Huang et~al.(2020)Huang, Xing, Dernoncourt, and Paul}]{huang-etal-2020-multilingual_mtc_hate_recog}
Huang, X.; Xing, L.; Dernoncourt, F.; and Paul, M.~J. 2020.
\newblock Multilingual {T}witter Corpus and Baselines for Evaluating Demographic Bias in Hate Speech Recognition.
\newblock In Calzolari, N.; B{\'e}chet, F.; Blache, P.; Choukri, K.; Cieri, C.; Declerck, T.; Goggi, S.; Isahara, H.; Maegaard, B.; Mariani, J.; Mazo, H.; Moreno, A.; Odijk, J.; and Piperidis, S., eds., \emph{Proceedings of the Twelfth Language Resources and Evaluation Conference}, 1440--1448. Marseille, France: European Language Resources Association.
\newblock ISBN 979-10-95546-34-4.

\bibitem[{Hutchinson and Mitchell(2019)}]{10.1145/3287560.3287600_50_years_fairness}
Hutchinson, B.; and Mitchell, M. 2019.
\newblock 50 Years of Test (Un)Fairness: Lessons for Machine Learning.
\newblock In \emph{Proceedings of the Conference on Fairness, Accountability, and Transparency}, FAT* '19, 49–58. New York, NY, USA: Association for Computing Machinery.
\newblock ISBN 9781450361255.

\bibitem[{Kapoor et~al.(2022)Kapoor, Dhawan, Goel, T~H, Bhatnagar, Agrawal, Agrawal, Bhattacharya, Kumaraguru, and Modi}]{kapoor-etal-2022-hldc}
Kapoor, A.; Dhawan, M.; Goel, A.; T~H, A.; Bhatnagar, A.; Agrawal, V.; Agrawal, A.; Bhattacharya, A.; Kumaraguru, P.; and Modi, A. 2022.
\newblock {HLDC}: {H}indi Legal Documents Corpus.
\newblock In Muresan, S.; Nakov, P.; and Villavicencio, A., eds., \emph{Findings of the Association for Computational Linguistics: ACL 2022}, 3521--3536. Dublin, Ireland: Association for Computational Linguistics.

\bibitem[{Kumar et~al.(2020)Kumar, Kumar, Kanojia, and Bhattacharyya}]{kumar-etal-2020-passage_hindi_fasttext}
Kumar, S.; Kumar, S.; Kanojia, D.; and Bhattacharyya, P. 2020.
\newblock {``}A Passage to {I}ndia{''}: Pre-trained Word Embeddings for {I}ndian Languages.
\newblock In Beermann, D.; Besacier, L.; Sakti, S.; and Soria, C., eds., \emph{Proceedings of the 1st Joint Workshop on Spoken Language Technologies for Under-resourced languages (SLTU) and Collaboration and Computing for Under-Resourced Languages (CCURL)}, 352--357. Marseille, France: European Language Resources association.
\newblock ISBN 979-10-95546-35-1.

\bibitem[{Landauer and Dumais(1997)}]{landauer1997solution_vector_averaging1}
Landauer, T.~K.; and Dumais, S.~T. 1997.
\newblock A solution to Plato's problem: The latent semantic analysis theory of acquisition, induction, and representation of knowledge.
\newblock \emph{Psychological review}, 104(2): 211.

\bibitem[{May et~al.(2019)May, Wang, Bordia, Bowman, and Rudinger}]{may-etal-2019-measuring_seat}
May, C.; Wang, A.; Bordia, S.; Bowman, S.~R.; and Rudinger, R. 2019.
\newblock On Measuring Social Biases in Sentence Encoders.
\newblock In Burstein, J.; Doran, C.; and Solorio, T., eds., \emph{Proceedings of the 2019 Conference of the North {A}merican Chapter of the Association for Computational Linguistics: Human Language Technologies, Volume 1 (Long and Short Papers)}, 622--628. Minneapolis, Minnesota: Association for Computational Linguistics.

\bibitem[{Prost, Thain, and Bolukbasi(2019)}]{prost-etal-2019-debiasing_bios}
Prost, F.; Thain, N.; and Bolukbasi, T. 2019.
\newblock Debiasing Embeddings for Reduced Gender Bias in Text Classification.
\newblock In Costa-juss{\`a}, M.~R.; Hardmeier, C.; Radford, W.; and Webster, K., eds., \emph{Proceedings of the First Workshop on Gender Bias in Natural Language Processing}, 69--75. Florence, Italy: Association for Computational Linguistics.

\bibitem[{Samadi et~al.(2018)Samadi, Tantipongpipat, Morgenstern, Singh, and Vempala}]{10.5555/3327546.3327755_price_of_fair_pca}
Samadi, S.; Tantipongpipat, U.; Morgenstern, J.; Singh, M.; and Vempala, S. 2018.
\newblock The Price of Fair PCA: One Extra Dimension.
\newblock In \emph{Proceedings of the 32nd International Conference on Neural Information Processing Systems}, NIPS'18, 10999–11010. Red Hook, NY, USA: Curran Associates Inc.

\bibitem[{Vu et~al.(2020)Vu, Nguyen, Le, and Jiang}]{vu-etal-2020-multimodal}
Vu, X.-S.; Nguyen, T.-S.; Le, D.-T.; and Jiang, L. 2020.
\newblock Multimodal Review Generation with Privacy and Fairness Awareness.
\newblock In Scott, D.; Bel, N.; and Zong, C., eds., \emph{Proceedings of the 28th International Conference on Computational Linguistics}, 414--425. Barcelona, Spain (Online): International Committee on Computational Linguistics.

\end{thebibliography}

\appendix
\section{Appendix}
\begin{figure*}
\begin{subfigure}{.49\textwidth}
    \centering
    \includegraphics[width=1\columnwidth]{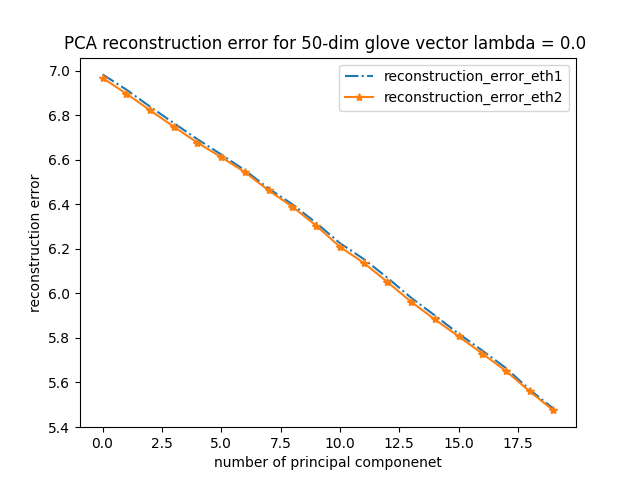}
    \caption{MTC Ethnicity Vector Extrema}
    \label{fig:mtce_ve}
\end{subfigure}
\begin{subfigure}{.49\textwidth}
    \centering
    \includegraphics[width=1\columnwidth]{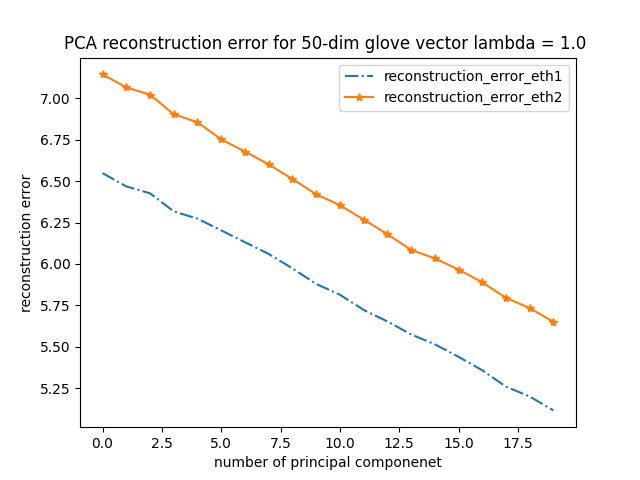}
    \caption{MTC Ethnicity Vector Averaging}
    \label{fig:mtce_va}
\end{subfigure}
    \caption{PCA reconstruction errors for each ethnicity group on the MTC-eth data}
    \label{fig:app-mtc-gen-pca-re}
\end{figure*}

The vector extrema PCA reconstruction errors are similar for both ethnicity groups across different number of dimensions used for reconstruction. However, vector average shows consistently higher reconstruction errors for ethnicity 2 over ethnicity 1 across dimensions. This can be clearly seen in the plots of Figure \ref{fig:app-mtc-gen-pca-re}

\begin{figure}
    \begin{subfigure}{.49\textwidth}
    \centering
    \includegraphics[width=1\columnwidth]{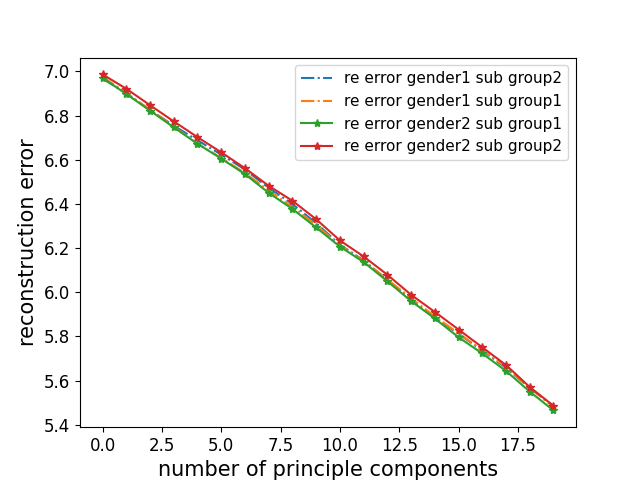}
    \caption{MTC-gen with vector extrema}
    \label{fig:mtcg_split_ve}
\end{subfigure}
    \caption{PCA reconstruction error within groups and across groups of MTC-gen using vector extrema encoding}
    \label{fig:pca_re_mtcg_ve}
\end{figure}

Figure \ref{fig:pca_re_mtcg_ve} shows the reconstruction errors of intra- and inter- groups of gender when using the vector extrema encoding. We find no discernible differences in the gaps between reconstruction errors of sub-groups and various groups.

\begin{figure}[!ht]
    \centering
    \includegraphics[width=1\columnwidth]{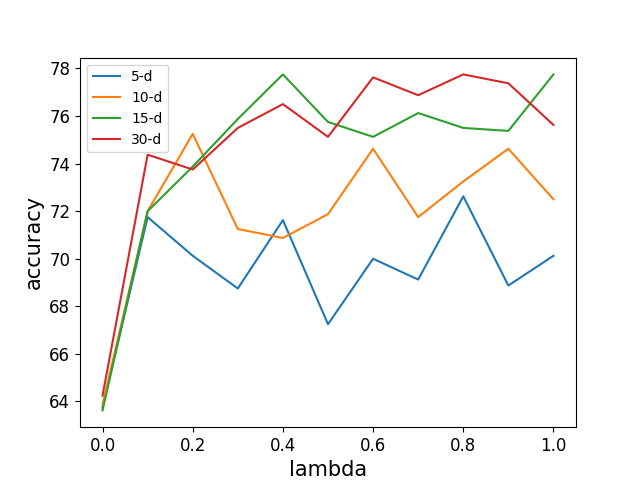}
    \caption{SVM accuracies on MTC-gen dataset}
    \label{fig:app_gen_svm2}
\end{figure}

Figure \ref{fig:app_gen_svm2} shows the accuracies of SVMs on MTC-gen dataset while using different number of dimensions (5d, 10d, 15d, 30d) of PCA reconstructed data with various values of $\lambda $ to form a convex combination of vector average and vector extrema encodings of the tweets.

\end{document}